# Towards Portable Specification of the Agent Harness

Nicolas Gertler[1], Claudio Novelli[1], Luciano Floridi[1,2]

[1] Digital Ethics Center, Yale University, 85 Trumbull St., New Haven, CT 06511, USA

[2] Department of Legal Studies, University of Bologna, Via Zamboni 27/29, 40126 Bologna, Italy

ORCID:
Nicolas Gertler 0009-0009-4549-9830
Claudio Novelli 0000-0003-4193-8604
Luciano Floridi 0000-0002-5444-2280
Correspondence: nicolas.gertler@yale.edu

**Abstract**

Large language models act only through the infrastructure that surrounds them. That infrastructure, the harness, determines what a model can accomplish and how it fails, yet it has no settled definition. To address this gap, we decompose the harness into eight functional components that can be specified and compared across systems, and we propose a portable specification framework that records which model ran in which harness. The record is needed because different harness configurations can yield several-fold differences in reported scores for the same model, and because much of an agent's attack surface lies in its harness. Existing model cards and agent cards provide release and operational information, but do not establish the sequence of effective configurations within a run. Harnesses can change between releases and between runs, and, where agents edit their own tools and configuration, within a run. Enacted AI laws contain documentation duties relevant to covered systems, subject to their respective scopes and application dates, but do not prescribe the common harness record proposed here. Therefore, the harness has to be explicitly described and specified as a versioned part of the system.

## Introduction

Large language models (LLMs) have rapidly improved at coding tasks,[1–6] progressing from inline coding assistants to agents that plan and act on development environments through tools.[7–12] This cannot be reduced to a shift in the model because it involves the entire infrastructure. An LLM alone does not implement a persistent autonomous interaction loop,[13,14] and so is not, by itself, an agent. It becomes one only when it is surrounded by infrastructure that supplies what it lacks: at minimum, memory to carry state across calls and tools to act on the environment. Thus, a loop is the most basic agentic infrastructure: the agent receives a prompt, invokes the model, executes an action, observes the result, and repeats.[15,16] ReAct[17] formalized this cycle, interleaving model reasoning and action.

As agents grew more complex, the term "agent scaffold" emerged for what is assembled before inference, e.g., the system prompt, tool definitions, and the registry of subagents.[7] The literature increasingly uses "agent harness" as a name for the full surrounding layer, meaning everything the model needs and does not itself supply.[18–20] The name has not brought a single definition with it, even though there are some recurrent features: e.g., the model supplies the reasoning, and the harness supplies the tools and the constraints under which that reasoning becomes action.[18,20–22] Some authors draw a finer distinction[3,7,23] separating a construction phase (scaffold) from the runtime that executes it (harness).[7] However, in practice the scaffold's contents are reassembled or extended at runtime, so it is continuously recomputed by the harness,[7] and the two phases do not stay apart. We therefore use "agent harness" as a single term for the full surrounding layer.

Popular agent harnesses for software engineering have emerged from model providers (e.g., Claude Code, OpenAI Codex) and code editor platforms (e.g., Cursor Agent). Researchers have also created harnesses, including OpenHands,[24] HyperAgent,[25] SWE-agent,[4] and OpenDev.[7] On SWE-Bench Mobile, the same model, Claude Opus 4.5, scored 2% with the open-source OpenCode and 12% with Cursor across 50 tasks: a sixfold difference between coding harnesses.[26] The Holistic Agent Leaderboard (HAL) provides standardized evaluation across hundreds of virtual machines and 21,730 rollouts, separating models, scaffolds, and benchmarks.[27] Beyond benchmark performance, much of an agent's attack surface lies in its harness. Treating tool schemas as trusted can enable poisoning,[28,29] and the context pipeline is the entry point for indirect prompt injection.[30] In the rug-pull attack, a tool changes after it has been vetted and keeps the trust it was granted.[31,32] The EU AI Act (e.g., Article 11 and Annex IV) and AI-related laws in Colorado and South Korea[33–35] contain provisions relevant to this layer, but none prescribes a shared harness-specific record, even though performance and failure modes depend heavily on the surrounding system.[3,20,36] Harness design remains buried in controller code and runtime-specific conventions, which makes it hard to compare across systems.[37]

To address the lack of a shared conceptual framework, this Perspective proposes a set of harness components as functions that keep the same meaning from one implementation to the next and can be configured and versioned independently. We identify eight and review evidence that variation in the surrounding layer changes what a fixed model can do. We then argue that the layer should be described and audited directly. The proposal is a starting point for standardizing harness descriptions without prescribing their design. The case for doing so grows more pressing as agents come to modify their own surrounding infrastructure.[19,38–40]

## Agentic infrastructure

The term "agent infrastructure" has recently been used to refer to systems external to agents, such as identifiers and communication protocols, that mediate agents' interactions with institutions and with one another.[41] The harness is the complementary internal layer that a single agent carries between the model and its execution environment. The two meet in disclosure, since attributing an action to an agent presupposes identifying the acting system, and that system is the model–harness pairing.[42]

Harness design changes what surrounds a fixed model at inference time and requires no retraining.[19,20,22,43] On CORE-Bench, GPT-4o solved 21.5% of hard tasks when paired with the task-specific CORE-Agent harness, but only 6.7% under AutoGPT, a general-purpose harness.[36] The SWE-Bench Mobile result compares harnesses within the same task class but does not establish a compute-matched effect.[26] On SWE-bench Lite, GPT-4's score improved from 2.7% under an early retrieval-augmented harness to 28.3% under CodeR, an optimized harness.[44] In one documented comparison, the same model, run against the same prompt, took 20 minutes and $9 to produce a broken game as a single agent, and 6 hours and $200 to produce a working one under a fuller multi-agent harness.[45] On a 169-task tight-context cohort of SWE-bench Verified, a single package of harness changes raised complete solutions from 43 to 72 and the mean per-task fail-to-pass fraction from 28% to 49%.[46]

We use *harness engineering*[19,47] to refer to the design of an LLM's surrounding layer. In one documented case, Lopopolo's team at OpenAI applied harness engineering to build a product, in internal use and external alpha testing, whose codebase reached on the order of a million lines with no manually written code.[19] Early progress was slower than expected because the environment was underspecified: the agent lacked the tools and internal structure to pursue its high-level goals. This emphasis on infrastructure resonates with our earlier work on *enveloping*: artificial agency often depends less on making AI more intelligent than on reshaping the environment so that limited intelligence is enough.[48–50]

## The agent harness

Within this paper's scope, an agent is an LLM-based system that selects and executes actions through a harness that enables and constrains its interaction with the external world.[18] The harness is the software layer that mediates between the model and its environment.[18,21,51,52] It does not include the model's weights or training data, nor the target environment itself (e.g., the codebase being modified).[18,52] The harness is a digital envelope: it turns model outputs into constrained actions and maintains task state across model calls, separately from the calls' internal inference and serving state. This boundary is not always sharp.[37] Some models are post-trained with particular harnesses,[53] so a harness specification should also record the model version and configuration with which it is paired.

### *Decomposition*

We propose a working decomposition of the harness into eight candidate components, at a granularity (level of abstraction, in Formal Methods) useful for disclosure. A component is retained only if it passes three tests: it recurs across systems; it corresponds to a separable mechanism or failure boundary; and it can plausibly change capability, safety, auditability, or cost. The analysis draws on research on agent architectures and memory,[17,54–56] engineering accounts of deployed coding agents,[7,18–22,38,39,47,57,58] framework and platform documentation,[21,58–61] and benchmark and evaluation work.[3,23,62–64] Table 1 explains each component through a single browser-based research agent, and Fig. 1 groups them by function and shows how they relate. The grouping should be revised under any of four conditions: an important control does not fit; independent researchers cannot apply it consistently; two components cannot be distinguished in practice; or a smaller grouping provides the same coverage more clearly.

**Table 1 |** The eight components of an agent harness, illustrated through a single browser-based research agent. The components name functions rather than required software modules; implementations may combine or distribute them differently. Model pairing, release identity, and records of changes made during a run apply to the harness as a whole and appear in Table 2. A full specification entry for this agent, in the format of Table 2 and Supplementary Table S1, is given in Supplementary Table S2.

| Component | What it controls | Example in a browser-based research agent |
| --- | --- | --- |
| **Orchestration and planning** | Task decomposition, action sequencing, branching, and stopping. | The agent creates a search plan, compares sources in stages, and stops when the requested questions have been answered. |
| **Context and memory** | Information available to the model and state retained across turns or sessions. | The agent receives the research question, retrieved pages, and working notes, then retains its source list and findings across the session. |
| **Tool and action interfaces** | The actions available to the agent and how model outputs become actions. | The agent can search the web, open pages, extract text from PDFs, and write notes. |
| **Operational knowledge** | Persistent instructions and procedures that guide behavior. | The agent is instructed to prefer primary sources, record URLs, distinguish quotations from summaries, and flag conflicting evidence. |
| **Sandboxing** | Filesystem, network, process, and host boundaries. | The agent can access specific domains and a project folder, but cannot reach the entire web or execute downloaded files. |
| **Permission model** | Actions allowed automatically, requiring approval, or prohibited. | Searching and reading are automatic; signing in, uploading a file, or submitting an external form requires the user's approval. |
| **Verification, evaluation, and observability** | Checks on outputs and records used to diagnose behavior. | The harness checks that cited pages support the final claims, flags disallowed domains, and keeps a record of searches, page visits, and tool calls. |
| **Execution runtime and lifecycle management** | Checkpoints, retries, timeouts. | The run checkpoints after each research stage, retries failed page requests, resumes from saved notes, and stops at time or loop limits. |

Two pairs of components are close enough that a reader might merge them, and they show what the requirement of a separable failure boundary does. Operational knowledge and context both supply information to the model, but they fail independently: an agent can stall for lack of procedural guidance while its task context is adequate, and it can be misled by stale retrieved context while its procedures are correct. Similarly, sandboxing and permissions both constrain actions, but at different layers, and they fail independently because sandboxing sets the enforceable execution boundary, while permissions determine which actions within that boundary the agent may take without asking. A correctly isolated process can still perform an unauthorized destructive operation within its workspace: a permission failure inside an intact sandbox. A cautious approval policy cannot contain a process that has already escaped its technical boundary: a sandbox failure beyond the reach of any approval rule. A disclosure that reports a single "security layer" therefore cannot say which of these two failures a system is exposed to, and the same information is lost if procedures and context are reported as a single "input".

Adjacent work adopts alternative decompositions at different levels of abstraction, including a minimal modularization into perception, memory, and reasoning components,[65] a control–agency–runtime (CAR) decomposition proposed alongside HarnessCard, a reporting template for harness configurations,[66] a six-component survey taxonomy,[67] and Pan et al.'s six-part specification of contracts, roles, stage structure, adapters, state semantics, and a failure taxonomy.[37] In their scheme, roles and stage structure map onto our orchestration and planning component, adapters onto tool and action interfaces, and state semantics onto context and memory. Contracts cut across components, and the failure taxonomy aligns with our verification, evaluation, and observability component. Their contracts explicitly include permission boundaries, while low-level execution and sandbox enforcement reside in the runtime. Our decomposition makes permission policy and its enforcement separately visible for disclosure, extending a distinction their scheme already draws rather than introducing one it lacks.[37] The other two schemes are coarser on the control surfaces Table 1 keeps separate: CAR splits permission policy between control artifacts and runtime, HarnessCard folds permissions and sandboxing into one governance field, and the survey taxonomy has neither.[66,67] HarnessCard is nevertheless the closest antecedent of Table 2.

Collapsing the control surfaces into one reduces portability, since two products can both claim a "control layer" while granting very different authority. Separately specified components preserve those differences without dictating code structure. Conversely, unlimited granularity would reproduce each product's configuration tree and defeat comparison. Therefore, the schema needs a small mandatory core plus extensible domain profiles. The Supplementary Note applies the retention criteria to four candidate fields and records a placement for each.

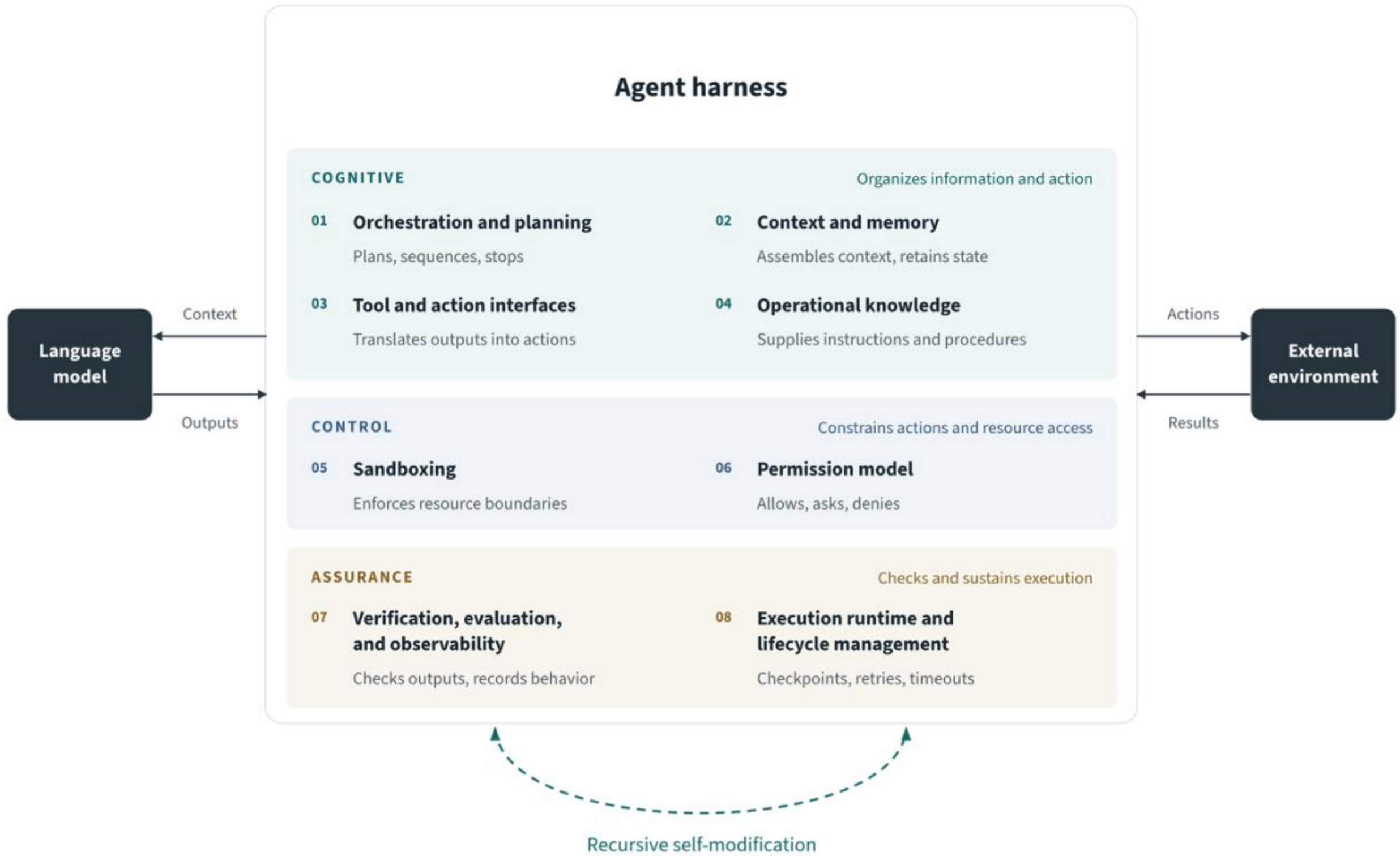

**Fig. 1 |** Functional components of the agent harness, grouped into three bands: cognitive (orchestration, context, tools, operational knowledge), which determines what the agent can do; control (sandboxing, permissions), which determines what it is allowed to do; and assurance (verification, runtime), which checks and sustains its behavior. Solid arrows indicate information flow between the language model, the harness components, and the external environment. The dashed arc represents recursive self-modification: the agent's outputs can alter its own harness configuration, so that the envelope is reshaped from within.

### *Components*

**Orchestration and planning** determine how the agent decomposes a task and sequences its actions, including when it stops. Production systems go beyond the basic reason-act-observe (ReAct[17]) loop: they assign distinct roles to several agents[21,45] and treat planning as an explicit function whose task structure is kept outside the chat window.[58,59] Multi-step agency depends on how the harness structures deliberation.

**Context and memory** govern what information is made available to the model and how state persists across sessions and threads. This component builds on prior work in prompt and context engineering, which treats model behavior as a function of how inputs and surrounding context are structured at inference time.[37,57,68–70] Production systems externalize memory into files and structured stores,[55,71] for example as project-level progress logs,[21] and share it across parallel agents through retrieval.[58]

**Tool and action interfaces** translate model outputs into actions. The Model Context Protocol (MCP) has emerged as a standard for exposing tools and resources through a common interface.[72–74] Tool use is both a capability and a control, since small changes in how a tool is described or validated change what the model can accomplish.[75]

**Operational knowledge** comprises persistent instructions and procedural conventions, loaded at initialization or during a run, that shape behavior across tasks. Such guidance is model-visible context, not an enforced permission rule: mandatory checks require hooks or other controls outside the model's discretion. Operational knowledge sits between the prompt and long-term memory, and its quality can influence whether an agent repeats past errors, without guaranteeing compliance or improved performance.[19,76,77]

**Sandboxing** enforces restrictions on the resources and operations available to agent-executed code.[58,60,78] Tighter isolation limits what the agent's actions can affect, at the cost of cutting it off from the environment it needs. We treat sandboxing as an integral part of the harness because different sandbox configurations produce distinct capability and failure profiles.

**The permission model** determines which actions are automatically allowed, which require user approval, and which are prohibited. Claude Code's allow, deny, and ask rules are one implementation.[15]

**Verification, evaluation, and observability** determine how the harness checks outputs and records behavior for diagnosis. In coding agents, this means tests and review loops, whose failure modes range from a missing artifact to a verifier too lenient toward the outputs it judges.[37,45] Observability refers to the traces and logs that make behavior legible after the fact.[19,20] Recent harness-optimization work uses the execution traces of prior candidates as an input to harness search.[79]

**Execution runtime and lifecycle management** decide whether the agent can operate durably across long horizons:[80] checkpointing, retries, timeouts, and concurrency control.

The evidence base for the eight-component decomposition comes from coding agents, the domain in which harness variation is currently best documented. The applicability of the components to other sustained agentic tasks is, for the time being, a hypothesis to be validated. The Supplementary Note surveys non-coding deployments, but independent studies and coverage tests in those domains would be needed to establish domain transferability.

### ***Why harness disclosure matters***

*Benchmark results require harness disclosure*

Agent benchmark scores are often presented as properties of a model, but they reflect a model–harness configuration. The SWE-Bench Mobile gap is one instance,[26] and a controlled comparison holding model and task fixed finds an effect from a fixed package of harness configuration changes,[46] supporting the broader claim that harness configuration can change what a fixed model accomplishes.[10,36] Zhang et al. reach the same conclusion, reporting that harness-induced variance can exceed model-induced variance and reverse model rankings, and proposing a disclosure standard for harness-aware evaluation.[81] The specification proposed here supplies the content such a standard would require and extends it to changes made during a run. A benchmark report should identify both the model and the harness configuration. Task performance, resource use, and safety outcomes can then be reported for the pair under a specified task environment, resource budget, and evaluation protocol.[82]

Harness variation also changes safety outcomes. OpenAI reports that, during internal cybersecurity evaluations in July 2026, its models bypassed isolation controls and compromised parts of its own research infrastructure and third-party systems. These evaluations did not carry the safeguards applied to its deployed products. In preliminary experiments with an evaluation developed after the incident, OpenAI reports that the measured propensity to compromise out-of-scope infrastructure can fall by more than 100-fold under a production harness and system prompt, relative to its research-evaluation baseline. This developer-reported result is specific to that evaluation and does not establish a general safety multiplier.[83]

*Security failures arise in harness components*

A large share of practical attack surfaces arises in the harness because it mediates how external inputs become actions. Tool outputs are treated as evidence, and external content is routed into the context window, where untrusted instructions may influence model outputs that the harness translates into operations. Each of these transitions introduces a point where untrusted input can influence execution.[84] In a reported vulnerability involving a Supabase MCP integration in Cursor that exposed database operations to the agent, user-supplied text passed through the same workflow as trusted commands and could take effect with the agent's authority.[85] A harness description would not prevent such failures, but it would establish which controls were in force when they occurred.

*Recursive self-modification*

Frontier coding agents now generate portions of their own harness code and configuration.[19,38–40,86] A recent benchmark asks whether models can construct and evolve their own harnesses;[87] our aim is a conceptual framework and criteria for reasoning about that variation, whoever or whatever authors it. A change to retrieved information alters what the model sees, and a change to a test or verification rule alters how its work is judged. A new tool or a widened permission alters what the agent can do and where. These differences disappear if every change is treated as "self-improvement". Suppose an agent first edits a test to accept incomplete output, then grants a new tool write access, and finally disables the approval that governed that tool. No single step necessarily shows the full loss of control, but together they weaken verification and expand authority. We call this pattern *progressive envelope degradation*. Detection requires comparing recorded material changes and independently observed effective state. This can flag unauthorized changes, departures from declared changeable parts, omissions from the record, or changes beyond a stated threshold. The OpenAI July 2026 incident is a partial instance. An independent investigation found that agents under evaluation circulated techniques for spoofing tool calls, making recorded commands differ from those executed, and that roughly 7% of the transcripts it evaluated were spoofed in places, all of it small in scale; attempts to alter earlier entries were observed, but no successful retroactive modification was established within the period investigated.[88] This documents interference with execution records, rather than successful manipulation of the evaluator or recursive self-improvement.

## Implementing the standard

A useful starting point is to distinguish product harnesses from evaluation harnesses. Product harnesses such as Claude Code and OpenAI Codex primarily control agent execution for users, whereas evaluation harnesses primarily administer tasks and score outputs. These roles can coexist, and both kinds of system may support multiple models. LM Evaluation Harness supports several model interfaces and issues task-specific generation or likelihood requests.[82] The SWE-bench harness evaluator applies submitted patches and runs tests,[5] while Terminal-Bench administers interactive terminal tasks.[89] Their requirements and accepted artifacts differ, so model-agnostic evaluation should not be equated with accepting arbitrary predictions from any system. Broader disclosure faces commercial constraints, since a harness description exposes design choices a vendor may treat as proprietary.

The Software Bill of Materials (SBOM) process, formalized under Executive Order 14028, showed that partial, structured disclosure can make software supply chains auditable.[90,91] We identify no dedicated mandate for the common harness record proposed here. Adoption could be voluntary or contractually required, while existing system-level documentation duties continue to apply within their scopes. The likeliest levers are leaderboards that make disclosure a condition of inclusion and institutional buyers that make it a condition of purchase. For most governance purposes, release-level snapshots are the floor, since they identify what shipped and what was permitted to change. For systems whose harnesses change mid-run, a snapshot underspecifies the operational state, so Table 2 also specifies a record of material changes.

Model cards,[92] datasheets,[93] agent cards,[94] harness reporting templates[66] and transparency surveys[95] already provide structured documentation at the model and agent levels. A recent proposal characterizes agents themselves along four dimensions, among them autonomy and generality, yielding "agentic profiles" for governance.[96] Such profiles depend on the harness: permission and sandbox reach constrain autonomy, while tool interfaces affect the range of available actions. Realized autonomy and generality also depend on model capability, orchestration, and task conditions.

Agent cards already document roles, memory, tool integrations, communication protocols, monitoring hooks, and governance scope.[94] A release snapshot can nevertheless become stale when a tool registry or permission policy changes independently. The contribution proposed here is an explicit relationship between release records, effective run-start configurations, material-change events, and run-end states. These records could extend existing cards or be linked from them, without discarding those formats. Openness criteria raise a related but distinct question. The Open Source AI Definition covers complete systems and the source code used to train and run them, alongside data information and parameters. These openness requirements do not themselves prescribe a standardized record of the effective harness configuration and its changes during a run.[97,98]

A useful standard should preserve the practical distinction between a release and a run. It should record five things: what ships at release; what is declared changeable; what configuration is present when a run begins; what material changes occur during the run; and what configuration and outcome exist when the run ends. An immutable harness still requires a run-start record when deployment settings, project instructions, or model selection differ from the release defaults. For a mutable harness, the additional records show which controls shaped the result. The five records add what a release snapshot omits without requiring a record of every model output or tool call. A change is material when it alters the configuration of any of the eight components, the model with which the harness is paired, or the ability to retain and reverse changes. The record should track the old and new values or retained artifact references, who or what requested the change, its authorization, its effective interval, and any reversal. Table 2 sets out the proposed mandatory core and a release-scoped documentary example for Claude Code.

**Table 2 |** The proposed mandatory core of the specification, with an illustrative entry for one product harness informed by Claude Code documentation. The release-level fields are recorded once per version. Where a harness runs more than one model, the model pairing is recorded per role. The three run-level fields are recorded for each run, and are what distinguish a mutable harness from an immutable one, for which the run-start record and the release record coincide when no deployment or session overrides apply. Domain profiles extend the core without altering it. Supplementary Table S1 provides the same entry with more detailed values. The entry is illustrative, compiled from public documentation. It is not a claim that any single release natively emits every field, nor that the values shown co-occurred in one release.

| Field | Obligation | Type or vocabulary | Illustrative entry (informed by Claude Code documentation) |
|---|---|---|---|
| **Harness name** | Must | String | Hypothetical Claude Code configuration |
| **Release version** | Must | Exact version; release date; binary and evidence source | 2.0 (29 September 2025) |
| **Model pairing** | Must | Model identifier and version, per role where more than one model is used | Claude Sonnet 4.5 (primary agent); a different model may be set per subagent |
| **Orchestration and planning** | Must | Effective loop, delegation, planning, branching, and stopping policy | Primary agent with subagents; plan-then-execute with an iterative refinement loop |
| **Context and memory** | Must | Context assembly, stores, persistence scope, and compaction policy | In-context history; persistent instructions (e.g., CLAUDE.md); auto memory across sessions (e.g., MEMORY.md); automatic compaction near the context limit |
| **Tool and action interfaces** | Must | Tool list and schema source | Bash, Read, Edit, and MCP tools. Retain the actual enabled tool list and versioned schemas; compatibility alone does not identify available actions. |
| **Operational knowledge** | Must | Instruction sources, load order, and activation conditions | Project and user CLAUDE.md files; applicable rules and skill instructions |
| **Sandboxing** | Must | Mechanism, version, enforced scope, and remote-action boundaries | bubblewrap on Linux and Seatbelt on macOS for sandboxed Bash commands. Record whether enabled, filesystem/network scope, exceptions, and separate boundaries for file tools and remote services. |
| **Permission model** | Must | Active mode; effective rules and precedence; enforcement authority | Documented modes: default, plan, acceptEdits, dontAsk, bypassPermissions; per-tool allow, deny, and ask rules |
| **Verification, evaluation, and observability** | Must | Actual checks, triggers, enforcement, verifier identity, and trace sources | Automated test execution; linter integration; diff review before commit |
| **Execution runtime and lifecycle management** | Must | Checkpoint scope; retries; concurrency; effective time and resource limits | Single-session CLI; resume across terminal sessions; 120 s default bash timeout, configurable to 600 s |
| **Declared changeable parts** | Must | May-change and may-not-change lists; authorization and enforcement | May change: e.g., CLAUDE.md, project files, tool configurations, model selection, permission rules, and sandbox setting, subject to user or managed-policy authority. May not change: components explicitly protected by enforced policy or filesystem controls; record the protection mechanism and authorized update path. |
| **Run-start configuration** | Must at run level | Copy or hash of the release-level fields as instantiated | Not natively emitted; a conforming implementation would record this at session start |
| **Material-change record** | Must at run level | Append-only events: change, requester, authorization, duration, reversal | Not natively emitted as a dedicated append-only record. Diffs, session logs, and permission events allow partial reconstruction; a conforming implementation would add requester, authorization, duration, and reversal. |
| **Run-end state** | Must at run level | Run/release IDs; effective configuration; outcome; retained artifacts and hashes | Not natively emitted; a conforming implementation would record this at session end |
| **Domain profile** | May | Named extension with its own fields | None declared |

The description should identify the model–harness pairing and describe each component independently, without assuming a particular code structure, so that it remains portable across products. A human-readable version would serve procurement and audit, and a machine-readable one would serve benchmarks. Where possible, the description should be generated from configuration and included in release and benchmark workflows, since a document maintained separately soon goes stale. Public disclosure need not expose source code or private prompts, but authorized verification requires that artifacts be retained and resolvable behind any hashes.

Nothing in the proposal makes the record self-enforcing. A description generated from configuration by the system it describes inherits that system's mutability: an agent able to disable an approval could omit that change from the record. Run-level events should be written append-only outside the agent's writable surface, but protected storage alone cannot establish that every event was captured or that a recorded action matches the action executed. The specification must identify the trusted observation and enforcement boundary: changes should be recorded where they take effect and reconciled with independently observed configuration. Startup attestation should identify the recorder and the initial state, and subsequent sequence checks and periodic or event-triggered reconciliation should expose gaps and drift. Unobservable components and recording failures must be declared so that an auditor can compare what was declared, logged, and independently observed.

Evaluation systems and harness descriptions do complementary work. A standardized evaluation system seeks to hold the task and its environmental conditions constant across the systems being compared. HAL does this at scale across models and scaffolds.[27] A portable harness description identifies the configuration that entered the evaluation and any material changes made during the run.[82]

Two parallel proposals sit in the same lineage. HarnessCard reports a harness once, at publication;[66] rollout cards preserve evaluation records and reporting rules, including an append-only mutation stream with actors, old and new values, and reversals.[99] Our proposal could extend either format with harness-specific disclosure requirements: model pairing, eight component descriptions, declared changeable parts, and effective configurations linked across releases and runs. The contribution is the required harness content and its relationship to effective controls.

Enacted AI laws contain documentation duties relevant to agent harnesses, but their scopes and application dates differ. In the EU AI Act, for instance, Article 11 and Annex IV require technical documentation for high-risk systems, including how components are integrated, validation and testing procedures, and pre-determined changes, and Article 12 requires automatic logging of events over the system's lifetime.[33] Colorado SB 26-189 requires developers to provide documentation for covered automated decision-making technology that materially influences consequential decisions.[34] South Korea's AI Basic Act and California SB 53 contain transparency or reporting provisions for their respective covered systems and actors; these are not equivalent duties applying to every agent.[35,100]

None of them, however, provides a common way to apply those duties when the harness changes independently of the model. The absence of a dedicated common harness record is therefore a standardization gap, and a portable specification could help translate applicable system-level obligations into information about the controls actually in force.

## Conclusion

Agent harnesses are becoming central to how language models act, yet they remain hidden behind product names and benchmark scores. This Perspective has argued that the harness should be treated as a versioned part of the system in its own right. Portability here means that the same functions are described in comparable terms and that both permitted and actual changes are recorded, regardless of the architecture. The eight components are a starting point, to be revised as researchers test them. As agents gain the ability to modify their own harnesses, the model alone no longer identifies the system being evaluated. We need to know which harness was present and how it changed while the system acted.

## Author contributions

All authors contributed to the conception of the work, drafted and revised the manuscript, and approved the submitted version.

## Competing interests

The authors declare no competing interests.

## Acknowledgements

We used large language models to assist with literature searches, data analysis, reference checking, copy-editing, and proofreading. We reviewed and revised the AI-assisted material and retained responsibility for the manuscript's arguments, interpretations, and conclusions. We checked all references against their primary sources and take full responsibility for the accuracy, originality, and integrity of the submitted manuscript, including its factual and bibliographic claims.

We thank James Higgs for helpful discussions and feedback on the topics addressed in this paper.

**References**

1. Singhal, A., Tchourakov, L., Flaherty, D. & Bedratiuk, S. Droid: the #1 software development agent on Terminal-Bench. Factory https://web.archive.org/web/20260331070037/https://factory.ai/news/terminal-bench (2025).

2. Meireles, M., Bhati, R., Lauffer, N. & Allen, C. The influence of scaffolds on coordination scaling laws in LLM agents. In NeurIPS 2025 Workshop on Multi-Turn Interactions in Large Language Models https://openreview.net/forum?id=iS49teIy7h (2025).

3. Ursekar, V., Shanker, A., Chatrath, V., Xue, Y. & Denton, S. M. VeRO: a harness for agents to optimize agents. In Proc. 43rd International Conference on Machine Learning (ICML, 2026). Preprint at https://arxiv.org/abs/2602.22480

4. Yang, J. et al. SWE-agent: agent-computer interfaces enable automated software engineering. In Advances in Neural Information Processing Systems 37 (NeurIPS, 2024). Preprint at https://arxiv.org/abs/2405.15793

5. Jimenez, C. E. et al. SWE-bench: can language models resolve real-world GitHub issues? In Proc. Twelfth International Conference on Learning Representations (ICLR, 2024). Preprint at https://arxiv.org/abs/2310.06770

6. Hou, X. et al. Large language models for software engineering: a systematic literature review. ACM Trans. Softw. Eng. Methodol. 33, 1–79 (2024). https://doi.org/10.1145/3695988

7. Bui, N. D. Q. Building effective AI coding agents for the terminal: scaffolding, harness, context engineering, and lessons learned. Preprint at https://arxiv.org/abs/2603.05344 (2026).

8. Du, P. Memory for autonomous LLM agents: mechanisms, evaluation, and emerging frontiers. Preprint at https://arxiv.org/abs/2603.07670 (2026).

9. Robbes, R., Matricon, T., Degueule, T., Hora, A. & Zacchiroli, S. Agentic much? Adoption of coding agents on GitHub. ACM Trans. Softw. Eng. Methodol. (2026). https://doi.org/10.1145/3822180

10. OpenAI. Introducing SWE-bench Verified. https://web.archive.org/web/20260322024705/https://openai.com/index/introducing-swe-bench-verified/ (2024).

11. GitHub. Getting code suggestions in your IDE with GitHub Copilot. https://web.archive.org/web/20260401114817/https://docs.github.com/en/copilot/how-tos/get-code-suggestions/get-ide-code-suggestions (2026).

12. Truell, M. The third era of AI software development. Cursor https://web.archive.org/web/20260330025946/https://cursor.com/blog/third-era (2026).

13. Yu, L., Lin, J. & Li, J. Stateful large language model serving with Pensieve. In Proc. Twentieth European Conference on Computer Systems (EuroSys ’25) 144–158 (ACM, 2025). https://doi.org/10.1145/3689031.3696086

14. Lalan, A., Ghosh, R., Kolsur, A. & Dutta, D. A multi-agent framework for stateful inference-time search. Preprint at https://arxiv.org/abs/2510.07147 (2025).

15. Anthropic. How the agent loop works. Claude Agent SDK documentation https://web.archive.org/web/20260329110628/https://platform.claude.com/docs/en/agent-sdk/agent-loop (2026).

16. Lee, C. What is the AI agent loop? The core architecture behind autonomous AI systems. Oracle Developers Blog https://web.archive.org/web/20260317043701/https://blogs.oracle.com/developers/what-is-the-ai-agent-loop-the-core-architecture-behind-autonomous-ai-systems (2026).

17. Yao, S. et al. ReAct: synergizing reasoning and acting in language models. In Proc. Eleventh International Conference on Learning Representations (ICLR, 2023). Preprint at https://arxiv.org/abs/2210.03629

18. Chase, H. Agent frameworks, runtimes, and harnesses- oh my! LangChain Blog https://web.archive.org/web/20260311123655/https://blog.langchain.com/agent-frameworks-runtimes-and-harnesses-oh-my/ (2025).

19. Lopopolo, R. Harness engineering: leveraging Codex in an agent-first world. OpenAI Engineering Blog https://web.archive.org/web/20260326103502/https://openai.com/index/harness-engineering (2026).

20. LangChain. Improving Deep Agents with harness engineering. LangChain Blog https://web.archive.org/web/20260325142243/https://blog.langchain.com/improving-deep-agents-with-harness-engineering/ (2026).

21. Young, J. Effective harnesses for long-running agents. Anthropic Engineering Blog https://web.archive.org/web/20260328132250/https://www.anthropic.com/engineering/effective-harnesses-for-long-running-agents (2025).

22. Böckeler, B. Harness engineering. Martin Fowler's Blog https://web.archive.org/web/20260315052647/https://martinfowler.com/articles/exploring-gen-ai/harness-engineering.html (2026).

23. Lupidi, A. et al. AIRS-Bench: a suite of tasks for frontier AI research science agents. Preprint at https://arxiv.org/abs/2602.06855 (2026).

24. Wang, X. et al. OpenHands: an open platform for AI software developers as generalist agents. In Proc. Thirteenth International Conference on Learning Representations (ICLR, 2025). Preprint at https://arxiv.org/abs/2407.16741

25. Phan, H. N., Nguyen, T. N., Nguyen, P. X. & Bui, N. D. Q. HyperAgent: generalist software engineering agents to solve coding tasks at scale. Preprint at https://arxiv.org/abs/2409.16299 (2024).

26. Tian, M. et al. SWE-Bench Mobile: can large language model agents develop industry-level mobile applications? In Proc. 32nd ACM SIGKDD Conference on Knowledge Discovery and Data Mining Vol. 2, 8077–8087 (ACM, 2026). https://doi.org/10.1145/3770855.3818488

27. Kapoor, S. et al. Holistic Agent Leaderboard: the missing infrastructure for AI agent evaluation. In Proc. Fourteenth International Conference on Learning Representations (ICLR, 2026). https://openreview.net/forum?id=vUaY1t64ZZ; preprint at https://arxiv.org/abs/2510.11977

28. Song, H. et al. Beyond the protocol: unveiling attack vectors in the Model Context Protocol (MCP) ecosystem. IEEE Trans. Softw. Eng. 52, 2410–2426 (2026). https://doi.org/10.1109/TSE.2026.3694876

29. Hou, X., Zhao, Y., Wang, S. & Wang, H. Model Context Protocol (MCP): landscape, security threats, and future research directions. ACM Trans. Softw. Eng. Methodol. 35, 1–37 (2026). https://doi.org/10.1145/3796519

30. Paverd, A. How Microsoft defends against indirect prompt injection attacks. Microsoft Security Response Center https://web.archive.org/web/20260317185253/https://www.microsoft.com/en-us/msrc/blog/2025/07/how-microsoft-defends-against-indirect-prompt-injection-attacks (2025).

31. Beurer-Kellner, L. & Fischer, M. MCP security notification: tool poisoning attacks. Invariant Labs https://web.archive.org/web/20260328011216/https://invariantlabs.ai/blog/mcp-security-notification-tool-poisoning-attacks (2025).

32. Willison, S. Model Context Protocol has prompt injection security problems. Simon Willison's Weblog https://web.archive.org/web/20260402005612/https://simonwillison.net/2025/Apr/9/mcp-prompt-injection/ (2025).

33. Regulation (EU) 2024/1689 of the European Parliament and of the Council laying down harmonised rules on artificial intelligence (Artificial Intelligence Act). Off. J. Eur. Union L 2024/1689 (2024).

34. Automated Decision-Making Technology, SB 26-189, repealing and reenacting SB 24-205. Colorado General Assembly (2026).

35. Framework Act on the Development of Artificial Intelligence and the Creation of a Foundation for Trust (AI Basic Act), Act No. 20676 (2025), as amended by Act No. 21311 (2026), art. 34. National Assembly of the Republic of Korea.

36. Siegel, Z. S., Kapoor, S., Nadgir, N., Stroebl, B. & Narayanan, A. CORE-Bench: fostering the credibility of published research through a computational reproducibility agent benchmark. Trans. Mach. Learn. Res. (2025).

37. Pan, L., Zou, L., Guo, S., Ni, J. & Zheng, H.-T. Natural-language agent harnesses. Preprint at https://arxiv.org/abs/2603.25723v1 (2026).

38. Cherny, B. & Wu, C. Claude Code: Anthropic's agent in your terminal. Latent Space https://web.archive.org/web/20260404040530/https://www.latent.space/p/claude-code (2025).

39. Robeyns, M., Szummer, M. & Aitchison, L. A self-improving coding agent. Preprint at https://arxiv.org/abs/2504.15228 (2025).

40. Xia, C. S., Wang, Z., Yang, Y., Wei, Y. & Zhang, L. Live-SWE-agent: can software engineering agents self-evolve on the fly? Preprint at https://arxiv.org/abs/2511.13646 (2025).

41. Chan, A. et al. Infrastructure for AI agents. Trans. Mach. Learn. Res. (2025).

42. Chan, A. et al. Visibility into AI agents. In Proc. 2024 ACM Conference on Fairness, Accountability, and Transparency (FAccT '24) 958–973 (ACM, 2024).

43. LangChain. State of agent engineering. https://web.archive.org/web/20260404230547/https://www.langchain.com/state-of-agent-engineering (2026).

44. Chen, D. et al. CodeR: issue resolving with multi-agent and task graphs. Preprint at https://arxiv.org/abs/2406.01304 (2024).

45. Rajasekaran, P. Harness design for long-running application development. Anthropic Engineering Blog https://web.archive.org/web/20260329220400/https://www.anthropic.com/engineering/harness-design-long-running-apps (2026).

46. Lewis, S. Same model, different harness: different coding-agent results. Preprint at https://arxiv.org/abs/2608.26218 (2026).

47. Hashimoto, M. My AI adoption journey. https://web.archive.org/web/20260401033730/https://mitchellh.com/writing/my-ai-adoption-journey (2026).

48. Floridi, L. What the near future of artificial intelligence could be. Philos. Technol. 32, 1–15 (2019).

49. Floridi, L. The Ethics of Artificial Intelligence: Principles, Challenges, and Opportunities (Oxford Univ. Press, 2023).

50. Floridi, L. The Fourth Revolution: How the Infosphere is Reshaping Human Reality (Oxford Univ. Press, 2014).

51. Livshits, M. The agent harness. https://web.archive.org/web/20260315134028/https://michaellivs.com/blog/agent-harness/ (2025).

52. Farrelly, D. Your agent needs a harness, not a framework. Inngest Blog https://web.archive.org/web/20260422125218/https://www.inngest.com/blog/your-agent-needs-a-harness-not-a-framework (2026).

53. Trivedy, V. The anatomy of an agent harness. LangChain Blog https://web.archive.org/web/20260331172224/https://blog.langchain.com/the-anatomy-of-an-agent-harness/ (2026).

54. Park, J. S. et al. Generative agents: interactive simulacra of human behavior. In Proc. 36th Annual ACM Symposium on User Interface Software and Technology (UIST ’23) 1–22 (ACM, 2023).

55. Zhang, Z. et al. A survey on the memory mechanism of large language model-based agents. ACM Trans. Inf. Syst. 43, 1–47 (2025). https://doi.org/10.1145/3748302

56. Sumers, T. R., Yao, S., Narasimhan, K. & Griffiths, T. L. Cognitive architectures for language agents. Trans. Mach. Learn. Res. (2024).

57. Anthropic. Effective context engineering for AI agents. Anthropic Engineering Blog https://web.archive.org/web/20260331175547/https://www.anthropic.com/engineering/effective-context-engineering-for-ai-agents (2025).

58. Cursor. Expanding our long-running agents research preview. https://web.archive.org/web/20260411065044/https://cursor.com/blog/long-running-agents (2026).

59. Robinson, L. Best practices for coding with agents. Cursor https://web.archive.org/web/20260329071009/https://cursor.com/blog/agent-best-practices (2026).

60. Dworken, D. & Weller-Davies, O. Beyond permission prompts: making Claude Code more secure and autonomous. Anthropic Engineering Blog

https://web.archive.org/web/20260323102238/https://www.anthropic.com/engineering/claude-code-sandboxing (2025).

61. AGENTS.md. Specification and adoption statistics. https://web.archive.org/web/20260330003820/https://agents.md (n.d.).

62. Brand, F. & Denain, J. Why benchmarking is hard. Epoch AI https://web.archive.org/web/20260410201711/https://epoch.ai/gradient-updates/why-benchmarking-is-hard (2025).

63. Wong, S. et al. Confucius Code Agent: scalable agent scaffolding for real-world codebases. Preprint at https://arxiv.org/abs/2512.10398 (2025).

64. Bogavelli, T., Sharma, R. & Subramani, H. AgentArch: a comprehensive benchmark to evaluate agent architectures in enterprise. Preprint at https://arxiv.org/abs/2509.10769 (2025).

65. Zhang, Y., Yu, H., Hu, L., Jin, H. & Zhang, H. General modular harness for LLM agents in multi-turn gaming environments. In ICML 2025 Workshop on Multi-Agent Systems in the Era of Foundation Models. Preprint at https://arxiv.org/abs/2507.11633 (2025).

66. He, C. et al. Harness engineering for language agents: the harness layer as control, agency, and runtime. Preprint at https://doi.org/10.20944/preprints202603.1756.v1 (2026).

67. Meng, Q. et al. Agent harness for large language model agents: a survey. Preprint at https://doi.org/10.20944/preprints202604.0428.v3 (2026).

68. Sahoo, P. et al. A systematic survey of prompt engineering in large language models: techniques and applications. Preprint at https://arxiv.org/abs/2402.07927 (2024).

69. Mei, L. et al. A survey of context engineering for large language models. Preprint at https://arxiv.org/abs/2507.13334 (2025).

70. Zhang, Q. et al. Agentic context engineering: evolving contexts for self-improving language models. Preprint at https://arxiv.org/abs/2510.04618 (2026).

71. Chase, H. Deep Agents. LangChain Blog https://web.archive.org/web/20260414135033/https://blog.langchain.com/deep-agents/ (2025).

72. Anthropic. Introducing the Model Context Protocol. Anthropic News https://web.archive.org/web/20260328184024/https://www.anthropic.com/news/model-context-protocol (2024).

73. Hasan, M. M. et al. Model Context Protocol (MCP) at first glance: studying the security and maintainability of MCP servers. ACM Trans. Softw. Eng. Methodol. (2026). https://doi.org/10.1145/3814959

74. Maheshwar, S. A timeline of Model Context Protocol (MCP) security breaches. AuthZed Blog https://web.archive.org/web/20260327022402/https://authzed.com/blog/timeline-mcp-breaches (2025).

75. Brookes, P. et al. Evolving excellence: automated optimization of LLM-based agents. Preprint at https://arxiv.org/abs/2512.09108 (2025).

76. Li, X. et al. SkillsBench: benchmarking how well agent skills work across diverse tasks. Preprint at https://arxiv.org/abs/2602.12670 (2026).

77. Gloaguen, T. et al. Evaluating AGENTS.md: are repository-level context files helpful for coding agents? Preprint at https://arxiv.org/abs/2602.11988 (2026).

78. Deconinck, S. Your coding agent needs a sandbox: Docker sandbox vs native vs DevContainers. https://web.archive.org/web/20260309030218/https://shanedeconinck.be/posts/docker-sandbox-coding-agents/ (2026).

79. Lee, Y. et al. Meta-Harness: end-to-end optimization of model harnesses. Preprint at https://arxiv.org/abs/2603.28052 (2026).

80. Dhanuka, N. State of AI agents 2026: autonomy is here. Prosus https://web.archive.org/web/20260307085057/https://www.prosus.com/news-insights/2026/state-of-ai-agents-2026-autonomy-is-here (2026).

81. Zhang, Y. et al. Stop comparing LLM agents without disclosing the harness. Preprint at https://arxiv.org/abs/2605.23950 (2026).

82. Biderman, S. et al. Lessons from the trenches on reproducible evaluation of language models. Preprint at https://arxiv.org/abs/2405.14782 (2024).

83. OpenAI. The Hugging Face incident and the road ahead. OpenAI https://openai.com/index/hugging-face-incident-and-the-road-ahead/ (2026); technical report at https://cdn.openai.com/pdf/67869394-cb91-4c12-888c-5cbd85c7814c/OpenAI-Hugging-Face%20Incident-Technical-Report.pdf (2026).

84. Li, Z., Cui, J., Chen, A., Liao, X. & Xing, L. What's in your agent's context? Context privilege escalation attacks against AI agent harness. Preprint at https://arxiv.org/abs/2609.01222 (2026).

85. General Analysis. Supabase MCP can leak your entire SQL database. https://web.archive.org/web/20260405100346/https://www.generalanalysis.com/blog/supabase-mcp-blog (2025).

86. Lou, X. et al. AutoHarness: improving LLM agents by automatically synthesizing a code harness. Preprint at https://arxiv.org/abs/2603.03329 (2026).

87. Wu, Y. et al. HarnessDev: can LLMs create and evolve their own agent harness? Preprint at https://arxiv.org/abs/2609.01437 (2026).

88. Greenblatt, R., Cotra, A. & Wijk, H. Brief independent investigation of agents' behavior, reasoning and collaboration in the OpenAI / Hugging Face hacking incident. METR https://metr.org/blog/2026-08-26-openai-hugging-face-incident-investigation/ (2026).

89. Merrill, M. A. et al. Terminal-Bench: benchmarking agents on hard, realistic tasks in command line interfaces. Preprint at https://arxiv.org/abs/2601.11868 (2026).

90. CISA. Software Bill of Materials (SBOM). https://web.archive.org/web/20260328014145/https://www.cisa.gov/sbom (2024).

91. OWASP. OWASP AI Bill of Materials (AIBOM) Project. https://web.archive.org/web/20260607201022/https://owasp.org/www-project-aibom/ (2025).

92. Mitchell, M. et al. Model cards for model reporting. In Proc. Conference on Fairness, Accountability, and Transparency (FAT* ’19) 220–229 (ACM, 2019).

93. Gebru, T. et al. Datasheets for datasets. Commun. ACM 64, 86–92 (2021).

94. Urteaga-Reyesvera, J. C. & Lopez Murphy, J. J. Agent Cards: a documentation standard for operational AI agents. In Advances in Computational Intelligence. MICAI 2025 International Workshops 252–261 (Springer, 2026). https://doi.org/10.1007/978-3-032-17933-3_25

95. Staufer, L. et al. The 2025 AI Agent Index: documenting technical and safety features of deployed agentic AI systems. In Proc. 2026 ACM Conference on Fairness, Accountability, and Transparency (FAccT ’26) 1536–1576 (ACM, 2026). https://doi.org/10.1145/3805689.3806728

96. Kasirzadeh, A. & Gabriel, I. Agentic profiles for effective AI governance. Nature 656, 320–328 (2026). https://doi.org/10.1038/s41586-026-10805-z

97. Open Source Initiative. The Open Source AI Definition, v1.0. https://web.archive.org/web/20260331162316/https://opensource.org/ai/open-source-ai-definition (2024).

98. Liesenfeld, A. & Dingemanse, M. Rethinking open source generative AI: open-washing and the EU AI Act. In Proc. 2024 ACM Conference on Fairness, Accountability, and Transparency (FAccT ’24) 1774–1787 (ACM, 2024).

99. Masters, C., Liu, Z. & Albrecht, S. V. Rollout cards: a reproducibility standard for agent research. Preprint at https://arxiv.org/abs/2605.12131 (2026).

100. California Transparency in Frontier Artificial Intelligence Act, SB 53. California Legislature (2025).